\documentclass{article}

\usepackage[final]{corl_2020} 

\usepackage{amsmath}
\usepackage{graphicx}
\usepackage{subfig}
\usepackage{amssymb}
\usepackage{float}
\usepackage{color}
\usepackage{bm}
\usepackage{tabularx}
\usepackage{mathtools}

\usepackage{amsmath,amsfonts,bm}

\def\eqref#1{equation~\ref{#1}}
\def\Eqref#1{Equation~\ref{#1}}

\def\1{\bm{1}}

\DeclareMathAlphabet{\mathsfit}{\encodingdefault}{\sfdefault}{m}{sl}
\SetMathAlphabet{\mathsfit}{bold}{\encodingdefault}{\sfdefault}{bx}{n}

\DeclareMathOperator*{\argmax}{arg\,max}
\DeclareMathOperator*{\argmin}{arg\,min}

\title{High Acceleration Reinforcement Learning for Real-World Juggling with Binary Rewards}

\author{
  Kai Ploeger$^*$, Michael Lutter\thanks{Equal contribution} \:, Jan Peters \\
  Computer Science Department, Technical University of Darmstadt \\
  \texttt{\{kai, michael, jan\}@robot-learning.de} \\
}

\begin{document}
\maketitle


\begin{abstract}
    Robots that can learn in the physical world will be important to enable robots to escape their stiff and pre-programmed movements.
    For dynamic high-acceleration tasks, such as juggling, learning in the real-world is particularly challenging as one must push the limits of the robot and its actuation without harming the system, amplifying the necessity of sample efficiency and safety for robot learning algorithms.
    In contrast to prior work which mainly focuses on the learning algorithm, we propose a learning system, that directly incorporates these requirements in the design of the policy representation, initialization, and optimization.
    We demonstrate that this system enables the high-speed Barrett WAM manipulator to learn juggling two balls from 56 minutes of experience with a binary reward signal. The final policy juggles continuously for up to 33 minutes or about 4500 repeated catches.
    The videos documenting the learning process and the evaluation can be found at \textbf{\href{https://sites.google.com/view/jugglingbot}{https://sites.google.com/view/jugglingbot}}
\end{abstract}


\vspace{-0.5em}
\section{Introduction} \vspace{-0.75em}
Robot learning is one promising approach to overcome the stiff and pre-programmed movements of current robots. When learning a task, the robot autonomously explores different movements and improves its behavior using scalar rewards. In recent years, research has focused a lot on improving task-agnostic deep reinforcement learning (DRL) algorithms by changing either the optimization \cite{schulman2017proximal, fujimoto2018addressing}, the simulation to use perturbed physics parameters~\cite{OpenAI18Cube, tobin2017domain}, or the task to gradually increase complexity~\cite{klink2019self}. While these approaches have propelled learning robots to very complex domains in simulation, ranging from full-body control of humanoids~\cite{Deepmind17Parkour} to control of dexterous hands~\cite{OpenAI19RubicsCube, Levine19Balls}, most of these approaches are not applicable to learn on physical systems as they neglect the intricate complexities of the real world. On the physical system, the learning is constrained to real-time and a single instance. Hence, the learning must not damage the robot with jerky actions and must be sample efficient.

Consider the high-acceleration task of juggling two balls next to each other with a single anthropomorphic manipulator. The manipulator is required to throw a ball upwards, move to the right, catch and throw the second ball and return to the left in time to catch the first ball. 
To sustain this cyclic juggling pattern, the robot must always throw the ball sufficiently vertical and maintain precise timing. Therefore, this task pushes the limits of the robot to achieve the required high accelerations (of up to $8$g), while maintaining precise control of the end-effector and the safety of the physical system. The task is not only inherently difficult to master\footnote[1]{For reference, the untrained human jugglers of the lab achieve 2 repeated catches and improve to about 20 repeated catches after a few hours of training.} but also requires learning on the physical system.
Real-world experience is required as simulation models, while good at simulating contact-free rigid-bodies, cannot represent the non-linear effects close to the torque limits of the actuators and the highly dynamic contacts between end-effector and balls. For the tendon driven Barrett WAM, rigid-body-simulators also cannot model the dominating cable dynamics at high accelerations. Hence, simulation-based solutions cannot be transferred for very dynamic tasks given our current simulators. For this specific task even robot to robot transfer between robots of the same model is not possible. The learned policy fails immediately when transferred to a different Barrett WAM. Therefore, the optimal policy depends on the exact robot instance and must be learned on the individual robot. 
The high accelerations amplify the safety and sample efficiency requirements for learning on the physical system as collisions at high velocities severely damage the system. Furthermore, the high acceleration movements induce a lot wear and tear and cannot be executed for days. Therefore, high acceleration tasks are an ideal problem to test the limitations of current learning algorithms on the physical system.
    
To emphasize important considerations for building a real-world learning system for high-acceleration tasks, we describe our robot learning setup to learn juggling and the taken system design considerations. We limit ourselves to \emph{existing} methods to focus solely on the requirements for the real-world application of current learning approaches. This is in contrast to many prior works, which mainly propose new policy representations or policy optimizations and demonstrate that the new algorithm can be applied to the physical system. Instead, we focus on a single challenging task and evaluate which existing representations and algorithms can be used to build a safe and sample efficient robot learning system. We also highlight limitations of learning approaches, which - given the current knowledge - we do not dare to apply to the physical system for robot juggling. Afterwards, we describe one approach to solve robot toss juggling with two balls and an anthropomorphic manipulator and validate the chosen approach in the real-world. The used approach - in our opinion - optimally combines engineering expertise and learning to obtain a reliable, safe, and sample efficient solution for this task. 

Our contribution is (1) the application of robot learning to the challenging task of single-arm toss juggling with two balls and (2) highlighting the challenges of applying a learning system in the physical world for a high-acceleration task. 
In the following, we first cover the prior work on robot juggling and learning dynamical tasks on the physical robot in Section~\ref{sec:related}. Afterwards, we compare different approaches to learn real-world toss juggling in Section~\ref{sec:options} and describe the implemented learning system in Section~\ref{sec:approach}. The experimental setup and results are presented in Section \ref{sec:env}.

\begin{figure*}[t]
    \centering
    \subfloat[]{\includegraphics[width=0.20\textwidth]{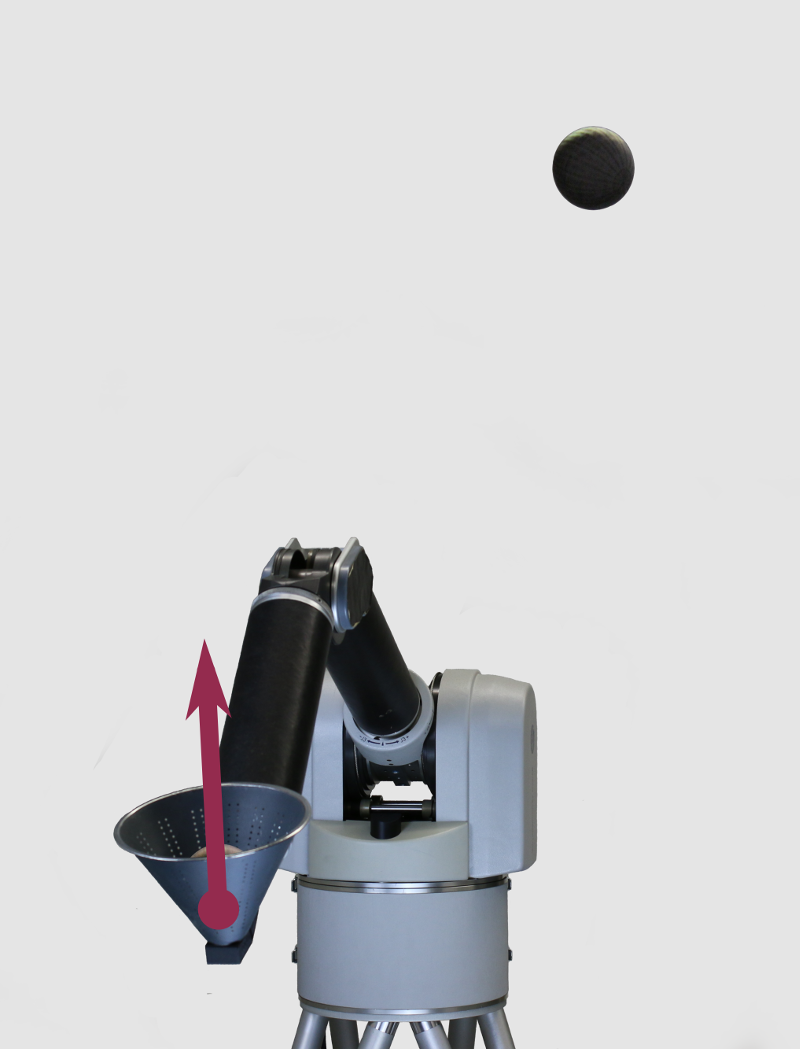}}
    \subfloat[]{\includegraphics[width=0.20\textwidth]{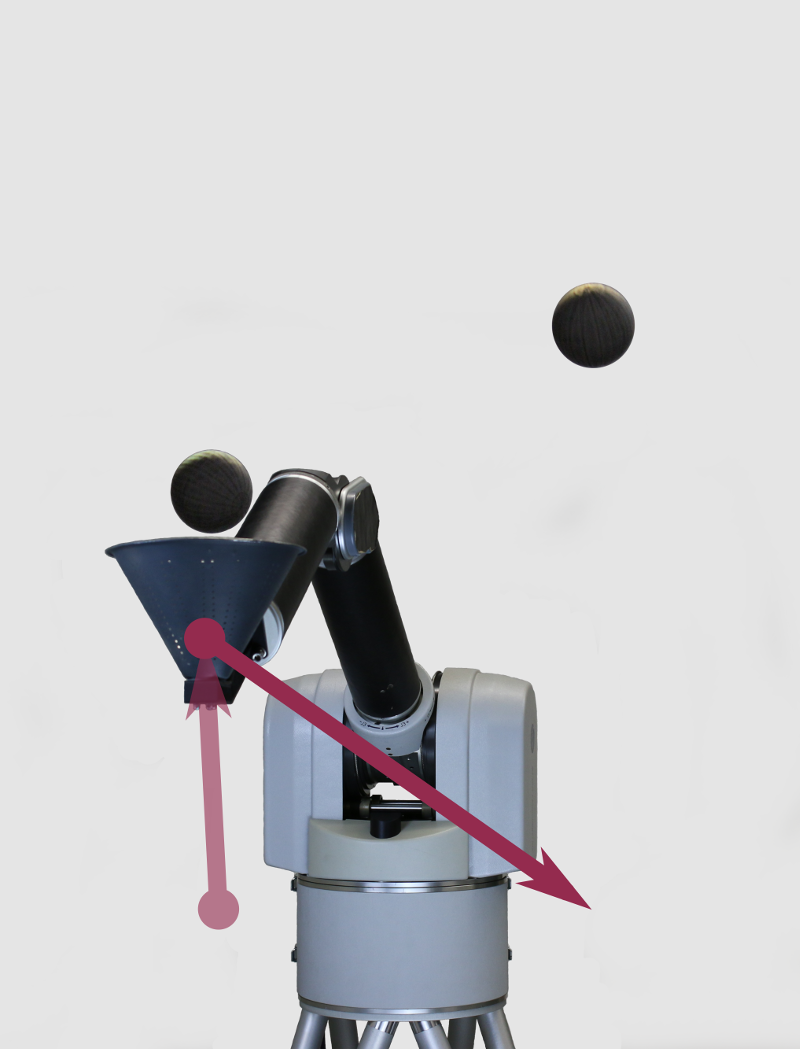}}
    \subfloat[]{\includegraphics[width=0.20\textwidth]{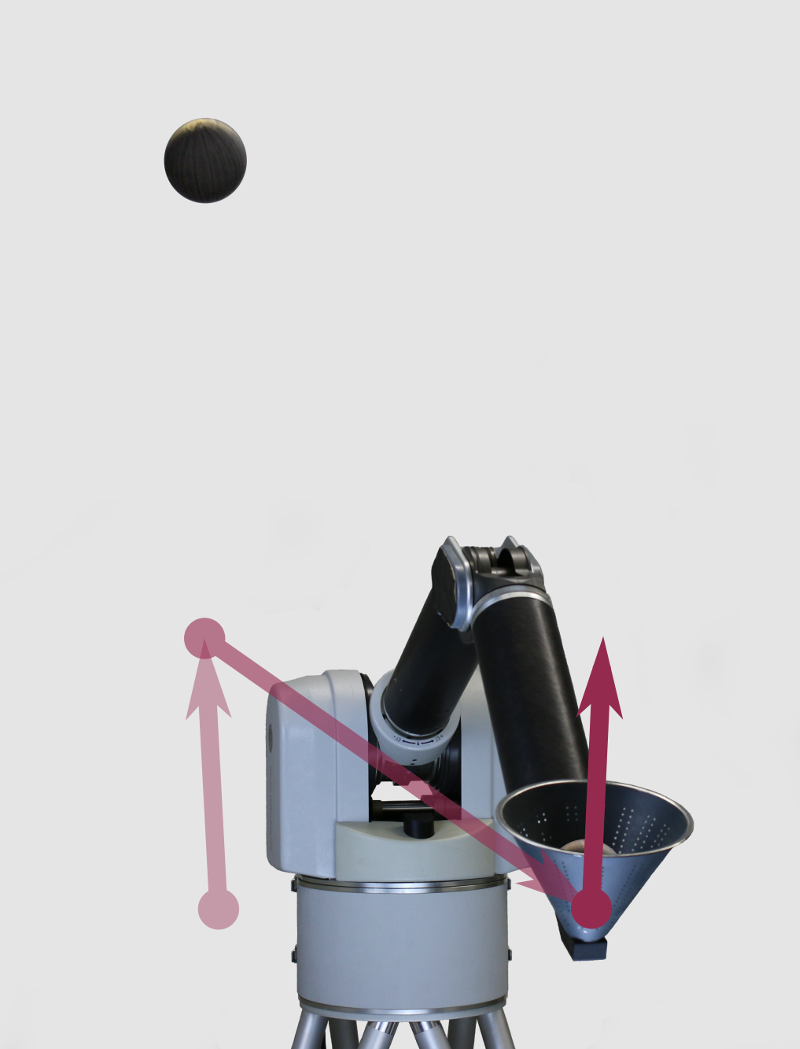}}
    \subfloat[]{\includegraphics[width=0.20\textwidth]{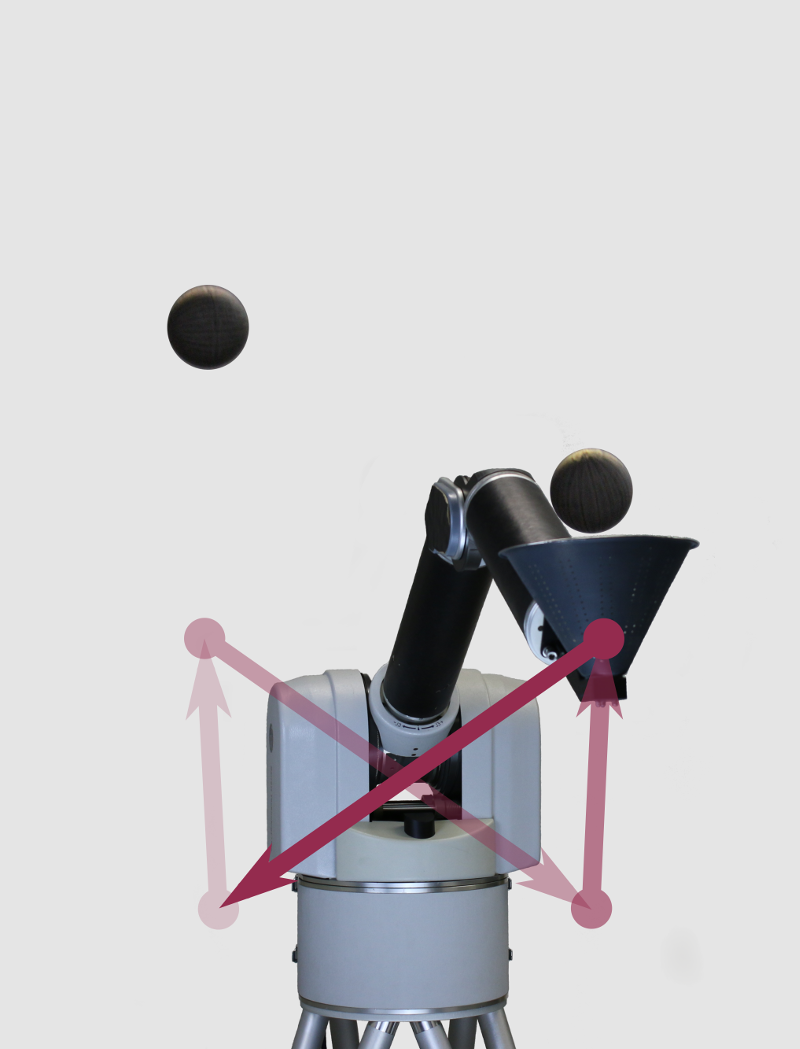}}
    \subfloat[]{\includegraphics[width=0.20\textwidth]{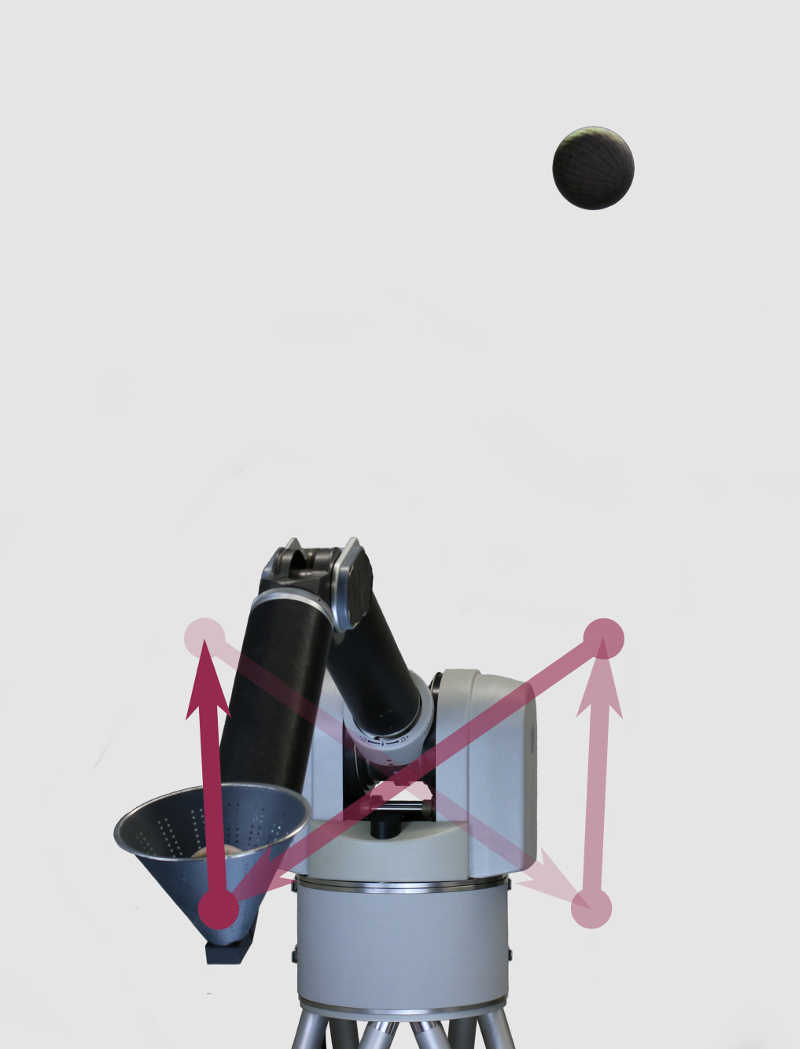}}
    \caption{The juggling movement consisting of four separate movements, which are repeated to achieve juggling of two balls with a single anthropomorphic manipulator: (a)~throw ball 1, (b)~catch ball 2, (c)~throw ball 2, (d)~catch ball one, (e)~repeat.}
    \label{fig:movement}
\end{figure*}


\vspace{-0.5em}
\section{Related Work} \label{sec:related}\vspace{-0.75em}

\subsection{Robot Juggling} \vspace{-0.75em}
For decades robot juggling has been used to showcase the ingenuity of mechanical system design and control engineering as listed in table~1. Starting with Claude Shannon in the 1970s, many different juggling machines were built.\footnote{A historic overview of robot juggling is available at \url{https://www.youtube.com/watch?v=2ZfaADDlH4w}.} For example, the Shannon juggler used cups to bounce up to five balls off a surface \cite{schaal1993open}, the devil-sticking machine stabilized a stick in the air \cite{Schaal_CSM_1994, schaal1994robot}, paddle jugglers, built from designated hardware \cite{reist2009bouncing, schaal1993open} or by attaching tennis rackets to manipulators \cite{aboaf1988task, aboaf1989task, rizzi1992distributed,rizzi1993further, Schaal_JMB_1996, 5509672, mueller2011quad}, juggled multiple balls using paddling. Toss jugglers were built using manipulators to juggle one \cite{174706, 292262} or two balls \cite{kizaki2012two} and even humanoids were used to juggle up to three balls using two arms \cite{riley2002robot, kober2012playing}. Most of these approaches proposed new engineering solutions for movements and controllers showing that these systems achieve juggling when the control parameters are manually fine-tuned. Only a few approaches used supervised learning for model learning \cite{aboaf1989task, schaal1994robot}, behavioral cloning \cite{5509672}, or evolutionary strategies with dense fitness functions \cite{Schaal_L_1993} to achieve paddle juggling and devil sticking. We build upon the vast experience on end-effector and controller design for robot juggling but in contrast to the prior work, we demonstrate, to the best of our knowledge, the first robot learning system that learns toss juggling with two balls and a single anthropomorphic manipulator in the real world using only binary rewards. Using this approach, we achieve juggling of up to 33 minutes and high repeatability between trials.

\begin{center}
    \footnotesize
    \label{tab:juggling}
    \captionof{table}{Prior work on different types of robot juggling}
    \begin{tabular}{| p{0.22\columnwidth} | p{0.34\columnwidth}  p{0.27\columnwidth} |} 
        \hline
        \textbf{Juggling Type} & \textbf{Approach} & \textbf{Papers} \\ 
        \hline
        Devil Sticking & Engineered & \cite{schaal1993open}\\  
        Devil Sticking & Model Learning & \cite{schaal1994robot} \\ 
        Paddle Juggling & Engineered & \cite{rizzi1992distributed, rizzi1993further, schaal1993open, Schaal_JMB_1996, reist2009bouncing, mueller2011quad}  \\ 
        Paddle Juggling & Imitation & \cite{5509672}  \\ 
        Paddle Juggling & Model Learning & \cite{aboaf1989task}  \\ 
        Paddle Juggling & Evolutionary Strategies & \cite{ Schaal_L_1993}  \\ 
        Toss Juggling & Engineered  & \cite{kizaki2012two, kober2012playing, riley2002robot, 292262, 174706}  \\ 
        \textbf{Toss Juggling}  & \textbf{Reinforcement Learning} & [\textbf{Ours}] \\ \hline
    \end{tabular}
\end{center}

\subsection{Learning Dynamical Tasks on the Physical Robot} \vspace{-0.75em}
Despite the recent surge of deep reinforcement learning algorithms for controlling robots, most of these approaches are constrained to learn in simulation due to sample complexity and the high risk of catastrophic policies. Only relatively few DRL approaches have been applied to physical robots, e.g., robot manipulators \cite{pinto2016supersizing, levine2018learning, Levine19Balls, johannink2019residual, kalashnikov2018qt, schwab2019simultaneously} or mobile robots \cite{kahn2020badgr, haarnoja2018soft}. Most of the work for manipulators focuses on non-dynamic tasks. Only \citet{schwab2019simultaneously} and \citet{buchler2020learning} applied DRL to learn the dynamic tasks of Ball-in-a-Cup or robot table tennis. In \cite{schwab2019simultaneously}, the authors built a fully automated environment to achieve large-scale data collection and engineered classical safety mechanisms to avoid damaging the physical system. Using the safe and automated environment, SAC-X was able to learn Ball-in-a-Cup from raw pixels within three days \cite{schwab2019simultaneously}. Most other approaches for learning dynamical tasks on the physical system use more classical robot learning techniques. These algorithms combine engineering- and task knowledge with learning to achieve sample efficient and safe learning that does not require completely safe and fully automated environments. For example, combining model learning with trajectory optimization/model predictive control \cite{NIPS2006_3151, Levine19Balls, schaal1994robot} or model-free reinforcement learning with engineered policy representation, expert policy initialization, and dense rewards \cite{5649089, kober2009policy, 5509672, kober2011reinforcement, mulling2013learning}. In our work, we extend the classical robot learning approach to a robot learning system that learns the high acceleration task of juggling with designed feature representations, expert policy initialization, and binary rewards instead of dense rewards. We also discuss the necessary design decisions that incorporate engineering and task expertise to achieve safety and sample efficiency. This focus is in contrast to most prior work as these mostly highlight the details of the learning algorithms but not the many engineering details that enable learning on the physical system.

\vspace{-0.5em}
\section{System Design for High-Acceleration Tasks} \label{sec:options}  \vspace{-0.75em}
Designing a robot learning system for robot juggling can be approached from different learning paradigms with different benefits. 
In the following we briefly discuss these trade-offs to motivate our approach presented in section~\ref{sec:approach}.

\subsection{Model-based vs. Model-free Learning} \vspace{-0.75em}
To minimize the wear and tear during the high acceleration movements and minimize the manual resets of picking up the balls, one desires to learn with minimal trials. 
Commonly model-based reinforcement learning (MBRL) methods are much more sample efficient compared to model-free reinforcement learning (MFRL) \cite{deisenroth2011pilco, chua2018deep}. However, MBRL requires to learn an accurate model describing the system such that the optimal policy transfers to the real system. For the considered task of robot juggling, the model would need to accurately describe the rigid-body dynamics of the manipulator, the stiction of the cable-drives, the contacts of the end-effectors with the balls, and the ball-movement.
The model would also need to be robust to out-of-distribution model exploitation to avoid optimizing a spurious and potentially harmful solution. Out-of-distribution exploitation is especially challenging for deep networks as the networks do not generalize well to previously unexplored states and this may result in unpredictable behaviors potentially damaging the system. 

We are not aware of a model-learning approach that can capture the different phenomena ranging from multi-point contacts to low-level stiction with sufficiently high fidelity and are robust to out of distribution generalization.
Therefore, we resort to a MFRL approach. Besides the theoretical aspects, the practical implementation of observing the necessary quantities to construct such accurate models is challenging by itself, e.g., the collisions of the ball and end-effector are obscured by the end-effector and hence not observable.

\subsection{Open-Loop Policy vs. Closed-Loop Policy} \vspace{-0.75em}
Closed-loop policies are favorable for robot juggling as the interactions between the end-effector and ball are not consistent. A closed-loop policy could recover from these irregularities.
Learning a closed-loop policy for juggling is non-trivial as one closes the loop on noisy ball observation. The noisy observations or potential outliers might cause the system to become unstable. The unstable behavior might cause the robot to hit its joint limits with high momentum and severely damage the system. One would need to guarantee that the closed-loop policy is stable for all possible observations including many edge-cases.
For the pre-dominant closed-loop deep network controller of deep reinforcement learning, the stable behavior cannot be guaranteed. Especially as the out of distribution prediction for networks is undefined. Currently, to the best of our knowledge, no network verification process for complex closed-loop systems exists. Therefore, we currently do not dare to apply a real-time closed-loop network policy to the physical system where the network is expected to execute accelerations of up to 8g.

Instead we are using an open-loop policy consisting of a desired position trajectory and a tracking controller. This representation is sufficient for the task as well trained jugglers can juggle basic patterns blindfolded and prior literature \cite{schaal1993open, schaal1994robot, Schaal_JMB_1996} has shown the stability of open-loop juggling robots.
A \emph{naive} but sufficient safety verification of this policy can be achieved by the stability of the tracking controller and by enforcing
tight box constraints in joint space for the desired trajectory. For the juggling setup the box-constraints prevent self-collisions and hitting the joint limits. 
Hybrid approaches that adapt the desired trajectory in closed-loop w.r.t. to ball observations exist. We also tried a \emph{naive} local adaption of the desired trajectory using a task-space controller but this adaptation even reduced system performance. The system would adapt to the balls during catching but could not throw straight afterwards.
As the open-loop policy already achieved high repeatability and long juggling duration, we did not investigate more complex hybrid policies further.

\vspace{-0.5em}
\section{System Implementation} \label{sec:approach} \vspace{-0.75em}
\subsection{Policy Representation} \label{sub:representation} \vspace{-0.75em}
The probabilistic policy is defined by a normal distribution $\mathcal{N}(\bm{\theta}; \bm{\mu}, \bm{\Sigma})$ over via-points in joint space. 
Each parameter  $\bm{\theta}=\left[\mathbf{q}_{0}, \dots \mathbf{q}_{N}, \dot{\mathbf{q}}_{0}, \dots, \dot{\mathbf{q}}_{N}, t_0, \dots, t_N \right]$ corresponds to a possible juggling movement consisting of a via-point sequence. Each via point is defined by the position $\mathbf{q}_i$, velocity $\dot{\mathbf{q}}_i$ and duration $t_i$. To execute the movement, the via-points are interpolated using cubic splines and tracked by a PD controller with gravity compensation. Therefore, the motor torques at each time-step are computed by,  
\begin{gather*}
    \bm{\tau} = \mathbf{K}_{\mathrm{P}}(\mathbf{q}_{\mathrm{ref}} - \mathbf{q}) + \mathbf{K}_{\mathrm{D}}(\dot{\mathbf{q}}_{\mathrm{ref}} - \dot{\mathbf{q}}) + \mathbf{g}(\mathbf{q})\\
    \text{with} \hspace{10pt} \mathbf{q}_{\mathrm{ref}}(t) = \sum_{j=0}^{3}\mathbf{a}_{i,j} \: (t-t_{i,0})^{j},  \hspace{10pt}
    \mathbf{\dot{q}}_{\mathrm{ref}}(t) = \sum_{j=1}^{3} j \: \mathbf{a}_{i,j} \: (t-t_{i, 0})^{j-1}, \hspace{10pt}
    t_{i,0} = \sum_{k=0}^{i-1} t_k
\end{gather*}
the control gains $\mathbf{K}_{\mathrm{P}}$ and $\mathbf{K}_{\mathrm{D}}$ and the $j$th parameter of the $i$th spline $\mathbf{a}_{ij}$. The gains are set to be low compared to industrial manipulators to achieve smooth movements. The spline parameters are computed using the via points by
\begin{gather*}
\mathbf{a}_{i,0} = \boldsymbol{q}_{i}, \hspace{15pt} \boldsymbol{a}_{i,1} = \mathbf{\dot{q}}_{i}, \hspace{15pt}
\mathbf{a}_{i, 2} = 3 \: \left(\boldsymbol{q}_{i+1} - \boldsymbol{q}_{i}\right)t_i^2 - \left(\boldsymbol{\dot{q}}_{i+1} + 2 \mathbf{\dot{q}}_{i}\right)t_i, \\ 
\mathbf{a}_{i, 3} = 2 \: \left(\boldsymbol{q}_{i} - \boldsymbol{q}_{i+1}\right)t_i^3 + \left(\boldsymbol{\dot{q}}_{i+1} + \mathbf{\dot{q}}_{i} \right)t_i^2.
\end{gather*}

We initialize the parameters with expert demonstrations to reduce the sample complexity. Especially in the case of binary rewards, such initialization is required as the reward signal is sparse. Most random initialization would not make any contact with the balls. Hence, the rl algorithm could not infer any information about the desired task. Instead of using kinesthetic demonstrations \cite{kober2009policy, 5649089, 5509672, kober2011reinforcement, mulling2013learning}, we directly initialize the interpretable via points manually. This initialization is preferable for robot juggling because the human demonstrator cannot achieve the necessary accelerations using kinesthetic teaching.

The desired juggling movement for two balls consists of four repeated movements, i.e., (a) throwing the first ball, (b) catching the second ball (c) throwing the second ball, and (d) catching the first ball (Fig. \ref{fig:movement}). We define the switching points between these movements as the via-points of the policy and to achieve a limit-cycle, we keep repeating these via-points. 
The cyclic pattern is prepended with an initial stroke movement that quickly enters the limit cycle without dropping a ball. Applying the limit cycles PD-references from the start would result in a bad first throw.
Furthermore, we enforce symmetry of cyclic pattern and zero velocity at the via point. Thus reducing the effective dimensionality from 48 open parameters to only 21.

\subsection{Policy Optimization}\label{sub:optimization} \vspace{-0.75em}
The policy optimization is framed as an episodic reinforcement learning problem, sampling a single policy parameter $\bm{\theta}_i \sim \mathcal{N}(\bm{\mu}, \bm{\Sigma})$ per roll-out and evaluating the episodic reward. This framing is identical to a bandit setting with high-dimensional and continuous actions. 
For the physical system, the episodic exploration is favorable over step-based exploration because this exploration yields smooth and stable action sequences given the previous policy representation. To optimize the policy parameters, we use a variant of the information-theoretic policy search approach episodic Relative Entropy Policy Search (eREPS) \cite{deisenroth2013survey, peters2010relative}. Our eREPS variation not only limits the Kullback-Leibler (KL) divergence when computing the sample weights but also enforces the reverse KL divergence when updating the policy parameters. This optimization can be solved in closed form for Gaussian distributions as described in \citet{abdolmaleki2017deriving}.
We use eREPS instead of EM-based~\cite{kober2009policy, 5649089} or gradient-based~\cite{kakade2002natural, peters2008natural} policy search as the KL constraint prevents premature convergence and large, possibly unsafe, jumps of the policy mean.\footnote{Further information about reinforcement learning for robotics can be found in the surveys \citet{kober2013reinforcement} and \citet{deisenroth2013survey}.}

Let the optimization problem of eREPS be defined as
\begin{align*}
    \pi_{k+1} = \argmax_{\pi} \int \pi(\bm{\theta}) R(\bm{\theta})d\bm{\theta},
    \hspace{15pt} \text{s.t.} \hspace{15pt} d_{\mathrm{KL}}(\pi_{k+1} || \pi_{k}) \leq \epsilon
\end{align*}
with the updated policy $\pi_{k+1}$, the episodic reward $R$, and the KL divergence $d_{\mathrm{KL}}$. With the additional constraint of $\pi$ being a probability distribution, this optimization problem can be solved by first optimizing the dual to obtain the optimal Lagrangian multiplier $\eta^{*}$ and fitting the new policy using weighted maximum likelihood. The sample-based optimization of the dual 
is described by 
\begin{align*}
\eta^* = \argmin_{\eta} \eta \epsilon + \eta \log \sum_{i=0}^{N} \pi_k(\bm{\theta}_i) \exp{\left(R(\bm{\theta}_i)/ \eta\right)}.
\end{align*}
The optimization of the weighted likelihood $\mathcal{L}$ to obtain the updated policy is described by
\begin{align*}
   \bm{\mu}_{k+1}, \: \bm{\Sigma}_{k+1} = \argmax_{(\bm{\mu}, \: \bm{\Sigma})} \sum_{i=0}^{N} w_i
   \log(\mathcal{L}(\bm{\mu}, \bm{\Sigma} | \bm{\theta}_i)) \hspace{15pt} \text{s.t.} \hspace{15pt}  d_{\mathrm{KL}}(\pi_k || \pi_{k+1}) \leq \epsilon.
\end{align*}
with the weights $w_i = \exp{\left(R(\bm{\theta}_i)/ \eta^{*}\right)} / \sum_i \exp{\left(R(\bm{\theta}_i)/ \eta^{*}\right)}$.
We incorporate the reverse KL constraint to the optimization to guarantee that the policy parameters adhere to the KL constraint. In the original eREPS formulation only the KL divergence between the evaluated samples is constrained but not the KL divergence between the parametrized policies. Especially for high-dimensional problems and few sample evaluations, the KL divergence between policies is larger than $\epsilon$ without this explicit constraint. The reverse KL divergence is used as this formulation enables solving the policy update in closed form for Gaussian distributions. For the multivariate Gaussian policy distribution this update is described by
\begin{gather*}
  \bm{\mu}_{k+1} = \frac{\xi^{*} \bm{\mu}_k + \bm{\mu}_s}{1 + \xi^{*}} ,  \hspace{20pt} 
  \bm{\Sigma}_{k+1} = \frac{\bm{\Sigma}_s + \xi^{*} \bm{\Sigma}_k + \xi^{*} \left(\bm{\mu}_{k+1} - \bm{\mu}_k \right)\left(\bm{\mu}_{k+1} - \bm{\mu}_k\right)^T}{1 +\eta^{*}} \\
  \bm{\mu}_s = \sum_{i=0}^{N} w_i \:  \bm{\theta}_i \hspace{20pt} 
  \bm{\Sigma}_s = \sum_{i=0}^{N} w_i \:  \left(\bm{\theta}_i - \bm{\mu}_{k+1}\right) \left(\bm{\theta}_i - \bm{\mu}_{k+1}\right)^T
\end{gather*}
with the optimal Lagrangian multiplier $\xi^{*}$. This optimal multiplier can be obtained by solving
\begin{align*}
    \xi^{*} = \argmax_{\xi} &\bigg[-\log(|\bm{\Sigma}_{k+1}|) - \sum_{i=0}^{N} w_i  \left(\bm{\theta}_i - \bm{\mu}_{k+1}\right)^T \bm{\Sigma}_{k+1}^{-1} \left(\bm{\theta}_i - \bm{\mu}_{k+1}\right) \bigg] 
    \\ 
     + \xi &\bigg[ 2\epsilon - \text{tr}\left( \bm{\Sigma}_{k+1}^{-1}  \bm{\Sigma}_{k} \right) - n + \log \left(\frac{ \bm{\Sigma}_{k+1}}{ \bm{\Sigma}_{k}} \right) + \left(\bm{\mu}_{k+1} - \bm{\mu}_k \right)^T \bm{\Sigma}_{k+1}^{-1} \left(\bm{\mu}_{k+1} - \bm{\mu}_k\right) \bigg].
\end{align*}
The complete derivation can be found in the appendix and the source code of eREPS is available at \href{https://github.com/hanyas/rl}{https://github.com/hanyas/rl}.

\begin{figure}[t]
    \centering
    \subfloat[]{\includegraphics[width=0.51\columnwidth]{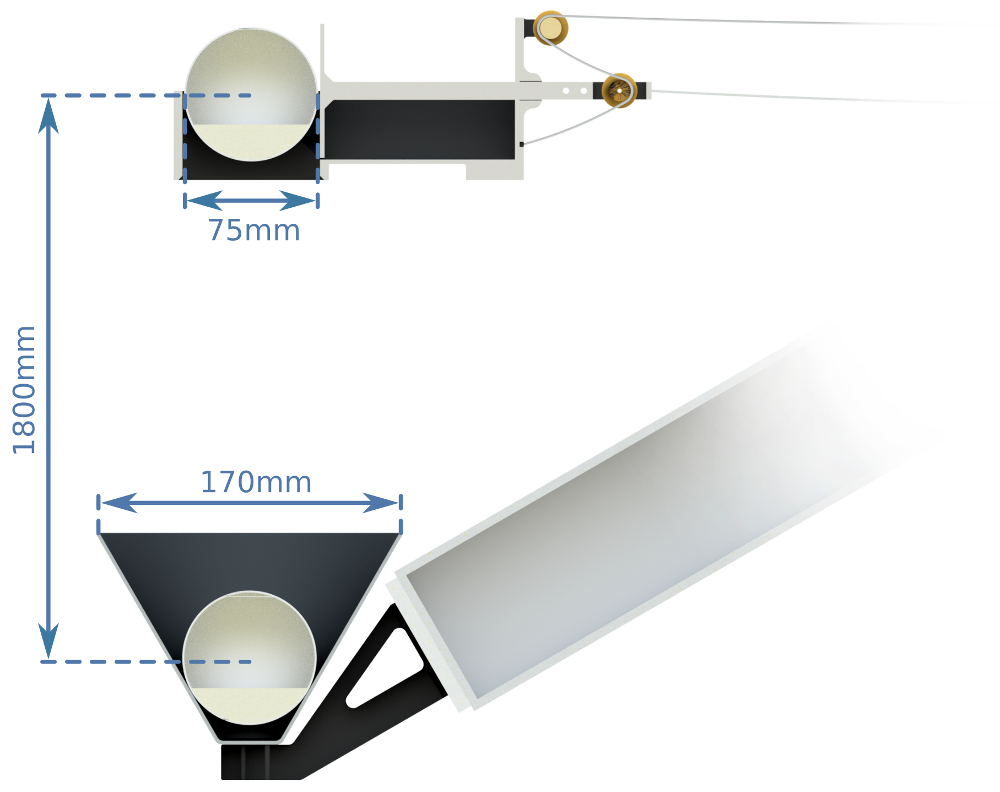} \label{fig:mechanics}}
    \subfloat[]{\includegraphics[width=0.49\columnwidth]{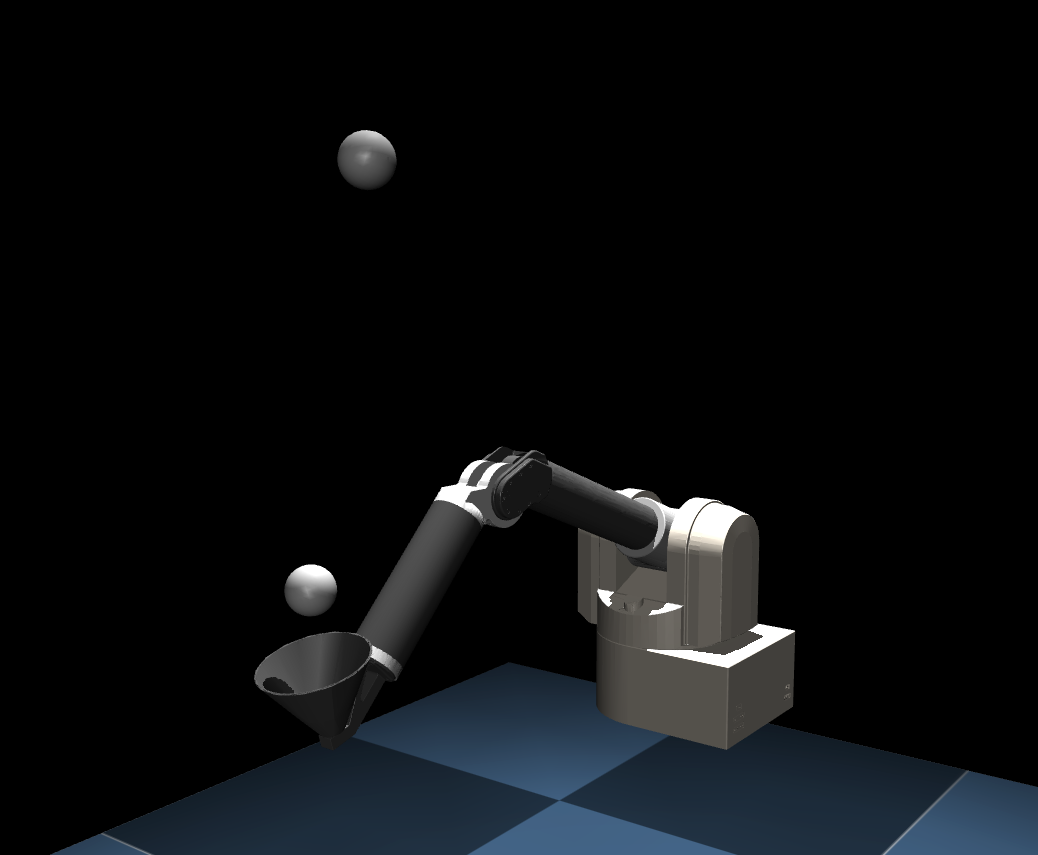} \label{fig:mujoco}}
    \caption{(a)~Mechanical design of the juggling end-effector, the Russian style juggling balls, and the ball launcher. The end-effector consists of a funnel to achieve precise control of the throwing movement. The Russian juggle balls are partially filled with sand to avoid bouncing.
    (b)~The MuJoCo juggling environment used for simulating the rigid-body-dynamics.}
\end{figure}

\subsection{Reward Function} \vspace{-0.75em}
The reward function assigns a positive binary reward as long as the robot is juggling. Juggling is defined as keeping both balls at least 60cm above the floor, which is measured using the external marker tracker OptiTrack. Therefore, the step-based reward is described by
\begin{equation*}
r(t) =
\left\{
	\begin{array}{ll}
		1  & \text{if } \min_{i} (b_{i, z}) \geq 0.6 \\
		0 & \text{otherwise} 
	\end{array}
\right.
\end{equation*}
with the $i$th ball height $b_{i,z}$. This reward signal maximizes the juggling duration and is not engineered to incorporate any knowledge about the desired ball- or manipulator trajectory. This choice of reward function is intuitive but also uninformative as a bad action only causes a delayed negative reward. For example, a bad action within the throwing will cause a zero reward - a drop - seconds after the action. This delay between action and reward, i.e., 'credit assignment', is a challenge in many RL problems.
The choice of the binary reward functions is in stark contrast to prior work as most of the previously proposed approaches use more informative dense rewards~\cite{schwab2019simultaneously, Levine19Balls, kober2009policy, Schaal_L_1993}. 

The binary rewards are favorable as these rewards do not require tuning a dense reward function. To specify the dense reward function one would need to predict the optimal ball trajectory and compute the distance to the optimal trajectory. Especially as the optimal ball trajectory depends on the robot capabilities and end-effector this prediction is challenging. Furthermore, one would need to align the initialization of the juggling movement with the optimal ball trajectory. One could possibly initialize a good juggling movement but that might not fit with the specified dense reward, e.g., the dense reward prefers higher throws than the initialization. With the binary reward one only needs to provide a good initialization and does not need to tune the reward parameters. Besides the reward tuning aspect, evaluating the dense reward is also more challenging compared to evaluating the binary reward in the real world. For evaluating the dense reward one would require precise tracking which is non-trivial to frequent occlusions by the end-effector. Even though we used OptiTrack to track the balls, we had a hard time achieving good tracking performance during the experiments due to the wear and tear on the reflective tape and the frequent occlusions. Every time a ball is in the end-effector, the ball is not observable. 

\vspace{-0.5em}
\section{Experiments} \label{sec:env}\vspace{-0.75em}

\subsection{Experimental Setup} \vspace{-0.75em}
The experiments are performed with a Barrett WAM to achieve the high accelerations required for robot juggling. The 4 degrees of freedom (DoF) variant is used rather than the 7 DoF version to allow for more precise control at high velocities and to save weight.
To achieve high repeatability, we use $75$mm Russian style juggling balls that dissipate the kinetic energy of the balls and prevent them from bouncing out of the end-effector. These balls consist of a hollow plastic shell partially filled with $37.5$g of sand (Figure~\ref{fig:mechanics}). The hard ball shells also result in more accurate throws compared to traditional beanbag juggling balls~\cite{292262, 174706}, as they do not deform.
As successfully described in prior toss juggling approaches~\cite{kober2012playing, riley2002robot, 292262, 174706}, the funnel-shaped end-effector passively compensates for minor differences in ball trajectories by sliding the balls from the edge of the $170$mm opening to the center.
The second ball is released via a launching mechanism $3$m above the floor, shown in Figure~\ref{fig:mechanics}, to achieve consistent height, velocity, and timing. This mechanism releases the ball by pushing the ball to an opening using a piston attached to a pulley system. The release of the ball is detected using Optitrack to start the episode.


\subsection{Simulation Studies} 
\begin{figure}[t]
    \centering
    \subfloat[]{\includegraphics[height=100pt]{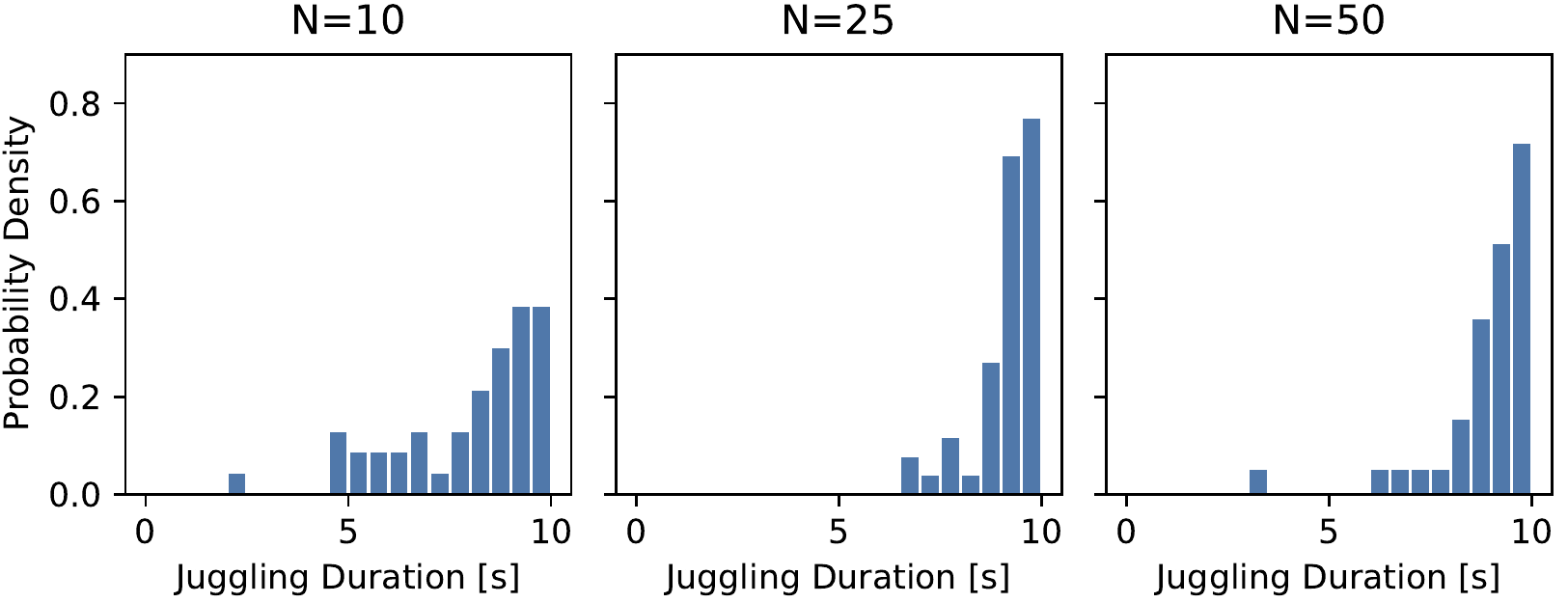} \label{fig:sim_hist}}
    \subfloat[]{\includegraphics[height=100pt]{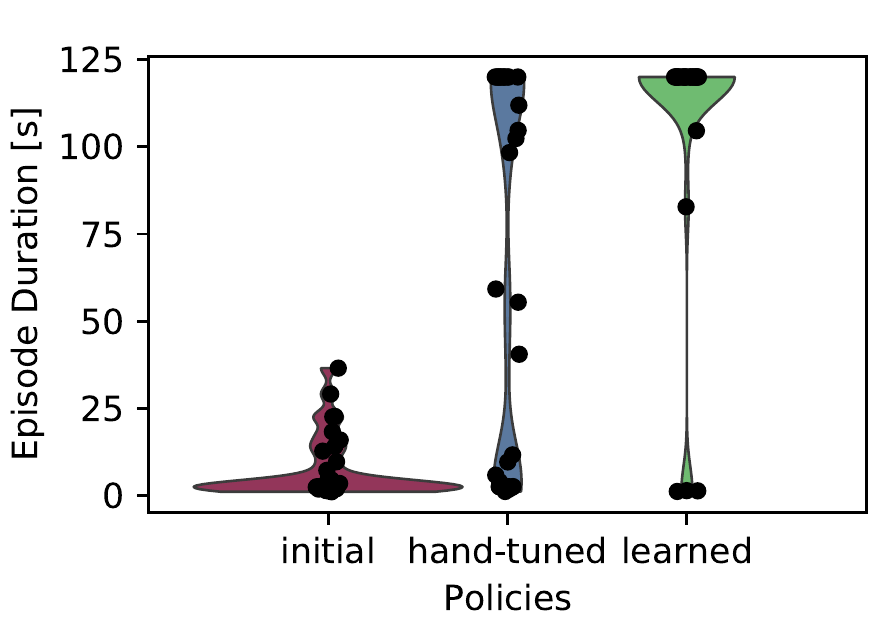} \label{fig:repeatability}}
    \caption{(a)~Mean juggling duration of final policies learned with varying batch sizes N. The maximal juggling duration is $10$s. (b)~Comparison of the learned and hand-tuned policy on each $30$ episodes with a maximum duration of $120$s on the real system. The learned policy achieves an average juggling duration of $106.82$s while the hand-tuned policy achieves $66.52$s.}
\end{figure}

\vspace{-0.75em}
The initial validation of the proposed learning system is performed in MuJoCo to evaluate the convergence of different numbers of roll-outs per episode over seeds. Figure \ref{fig:sim_hist} shows the juggling duration distribution of the final policy averaged over 60 different seeds at $10$, $25$ and $50$ roll-outs per episode. For $10$ roll-outs per episode, the learning system frequently converges to a sub-optimal final policy, which does not achieve consistent juggling of $10$ seconds. The probability of converging to a good policy is the highest for $25$ roll-outs per episode and hence, we use $25$ roll-outs per episode on the physical system. It is important to point out, that learning juggling in the simulation is actually more challenging compared to the real world as the stabilizing passive dynamics of the system could not be modeled accurately.
    
\begin{figure*}[h]
    \centering
    \includegraphics[width=\textwidth]{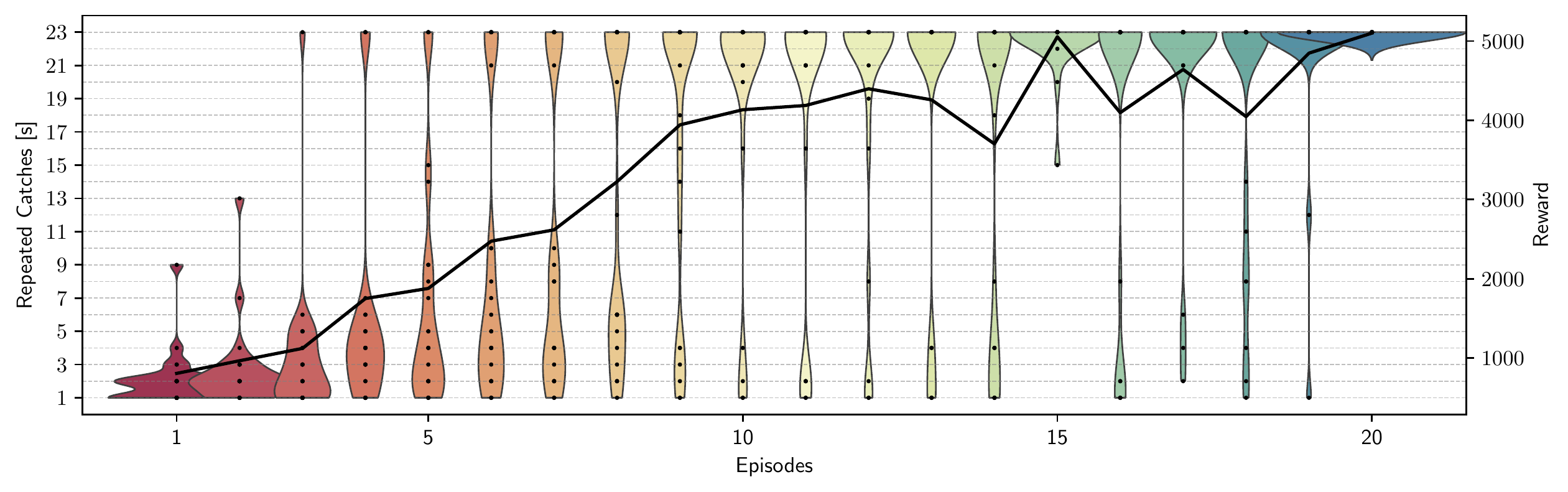}
    \caption{Reward distributions while learning to juggle on the physical system. Each point corresponds to a single roll-out on the system. Starting from the expert initialization, which only achieves juggling for a few repeated catches, the robot continuously improves using only binary rewards. After $56$ minutes of training, the final policy achieves consistent juggling for more than $10$s.}
    \label{fig:learning_curve}
    \vspace{-1.75em}
\end{figure*} 

\subsection{Learning on the real Barrett WAM} \vspace{-0.75em}
For the learning on the physical Barrett WAM $20$ episodes were performed. During each episode $25$ randomly sampled parameters were executed and the episodic reward evaluated. If the robot successfully juggles for $10$s, the roll-out is stopped. Roll-outs that were corrupted due to obvious environment errors were repeated using the same parameters. Minor variations caused by the environment initialization were not repeated. After collecting the samples, the policy was updated using eREPS with a KL constraint of $2$. The learning progress is shown in Figure \ref{fig:learning_curve}. Initially, the robot on average achieves between $1$ to $3$ repeated catches. These initial catches are important as the robot would otherwise not receive any meaningful information. Therefore, the rudimentary expert initialization must achieve some repeated catches to ensure fast learning. After the initial episodes, the robot rapidly improves the juggling duration. Starting from episode 10, the robot achieves consistent juggling of $10$ seconds and only very few balls are dropped during the start. During the next $10$ episodes, the average reward oscillates as the number of dropped balls varies but the robot achieves successful completion for the other trials. At episode $20$ the robot achieves perfect juggling of all $25$ randomly sampled parameters. Videos documenting the learning process and the evaluation can be found at \textbf{\href{https://sites.google.com/view/jugglingbot}{https://sites.google.com/view/jugglingbot}}. 

To test the repeatability and stability of the learned policy, the deterministic policy mean of episode $20$ is executed for $30$ repeated roll-outs with a maximum duration of $120$ seconds. The achieved performance is compared to a hand-tuned policy in Figure~\ref{fig:repeatability}.
Averaging at $106.82$s, the learned policy performs significantly better compared to the hand-tuned policy with $66.51$s. The weaker performance of the hand-tuned policy within the stroke-based initiation movement matches our expectations as tuning the stroke-based movement is the hardest part of the manual parameter tuning. Both policies do not achieve perfect repeatability due to the residual stochasticity of the environment.

To test the stability of the learned policy, the juggling was repeatedly executed and the maximum juggling duration recorded.
The learned policy achieved juggling for $33$ minutes, which corresponds to more than $4500$ repeated catches on the second try, after $15$ minutes on the first one. The high number of repeated catches, highlights the precision of the Barrett WAM, the end-effector design and both policies. Once the juggling is initiated successfully, the policies can recover from minor variations due to the passive stability induced by the end-effector design.


\vspace{-0.5em}
\section{Conclusion} \label{sec:conclusion} \vspace{-0.75em}
We described a robot learning system capable of learning toss juggling of two balls with a single anthropomorphic manipulator using only binary rewards. We demonstrated that our system can learn this high acceleration task within $56$ minutes of experience, utilizing sufficient engineering and task knowledge designing the robot learning system. Starting from a rudimentary expert initialization, the system consistently improves until achieving repeated juggling of up to $33$ minutes, which corresponds to more than $4500$ repeated catches. Furthermore, the learned policy outperforms a hand-tuned policy in terms of repeatability and achieves significantly higher rewards average across $30$ trials. In addition, we highlighted and discussed the incorporated engineering and task expertise to make learning on the physical system viable. This discussion should help future scientists and practitioners to address the needs of a physical system when building future robot learning systems. Nevertheless, this approach also pointed out the shortcomings of state-of-the-art robot learning approaches for learning dynamic tasks on the physical system. Despite the incorporated engineering and task knowledge the learning still takes up to $5$ hours and hence, reiterates the necessity for more sample efficient representations and learning approaches for sparse and binary rewards.

\newpage
\section*{Acknowledgments} \vspace{-0.75em}
This project has received funding from the European Union’s Horizon 2020 research and innovation program under grant agreement No \#640554 (SKILLS4ROBOTS). Furthermore, this research was also supported by grants from ABB and NVIDIA.

\bibliography{references}

\appendix

\newpage
\section{KL Constrained Maximum Likelihood Optimization for eREPS}
The standard eREPS formulation \cite{deisenroth2013survey} solves the optimization problem described by
\begin{align} \label{eq:ereps}
    \pi_{k+1} = \argmax_{\pi} \int \pi(\bm{\theta}) R(\bm{\theta})d\bm{\theta},
    \hspace{15pt} \text{s.t.} \hspace{15pt} d_{\mathrm{KL}}(\pi_{k+1} || \pi_{k}) \leq \epsilon
\end{align}
via computing the importance weights of each sample and fitting the policy to the weighted samples. The sample weights are described by 
\begin{align*}
w_i = \frac{\exp(R(\bm{\theta}_i)/ \eta^{*})}{\sum_i \exp(R(\bm{\theta}_i)/ \eta^{*})} \hspace{20pt} \text{with} \hspace{20pt} \eta^* = \argmin_{\eta} \eta \epsilon + \eta \log \sum_{i=0}^{N} \pi_k(\bm{\theta}_i) \exp{\left(R(\bm{\theta}_i)/ \eta\right)}.
\end{align*}
The fitting of the policy to the weighted samples is described by
\begin{align*}
   \bm{\mu}_{k+1}, \: \bm{\Sigma}_{k+1} = \argmax_{(\bm{\mu}, \: \bm{\Sigma})} \sum_{i=0}^{N} w_i
   \log(\mathcal{L}(\bm{\mu}, \bm{\Sigma} | \bm{\theta}_i)).
\end{align*}
In the case of Gaussian policies this optimization can be solved in closed form and the updated parameters are described by 
\begin{align*}
  \bm{\mu}_{k+1} = \sum_{i=0}^{N} w_i \:  \bm{\theta}_i \hspace{20pt} 
  \bm{\Sigma}_{k+1} = \sum_{i=0}^{N} w_i \:  \left(\bm{\theta}_i - \bm{\mu}_{k+1}\right) \left(\bm{\theta}_i - \bm{\mu}_{k+1}\right)^T.
\end{align*}
For an in-depth derivation of this approach please refer to \cite{deisenroth2013survey}. This approach to updating the policy works well if $N \gg n$ with the parameter dimensionality $n$. In this case the KL divergence between two consecutive parametrized policies is smaller than $\epsilon$. If $N \approx n$ the KL divergence between two consecutive parametrized policies must not necessarily be smaller than $\epsilon$. In this case the KL divergence between the weighted and unweighted samples is bounded by $\epsilon$ but the KL divergence between the parametrized policies is not. To ensure that the KL divergence between parametrized policies is bounded after the maximum likelihood optimization, we change the optimization problem to include a KL constraint. The constrained objective is described by 
\begin{align*}
   \bm{\mu}_{k+1}, \: \bm{\Sigma}_{k+1} = \argmax_{(\bm{\mu}, \: \bm{\Sigma})} \sum_{i=0}^{N} w_i
   \log(\mathcal{L}(\bm{\mu}, \bm{\Sigma} | \bm{\theta}_i)) \hspace{15pt} \text{s.t.} \hspace{15pt}  d_{\mathrm{KL}}(\pi_k || \pi_{k+1}) \leq \epsilon.
\end{align*}
Please note that the order of the KL divergence is switched compared to \Eqref{eq:ereps} and that the KL divergence is not symmetric. We switch from the I-projection to the M-projection because otherwise this optimization has no closed form solution for Gaussian policies. For bounding the distance between two consecutive policies the reverse KL divergence can be used as for small KL divergences both projections are comparable. The constrained optimization problem can be solved using Lagrangian multipliers as initially derived by \cite{abdolmaleki2017deriving}. The Lagrangian $L$ for a Gaussian policy is described by
\begin{align*}
    L =&\sum_{i=0}^{N} w_i
   \log(\mathcal{L}(\bm{\mu}, \bm{\Sigma} | \bm{\theta}_i)) + \xi (\epsilon - d_{\mathrm{KL}}(\mathcal{N}(\bm{\mu}_{k}, \: \bm{\Sigma}_{k}) \: || \: \mathcal{N}(\bm{\mu}_{k+1}, \: \bm{\Sigma}_{k+1} ))) \\
   = &\frac{1}{2}\bigg[- n \log(2\pi) - \log(|\bm{\Sigma}_{k+1}|) - \sum_{i=0}^{N} w_i  \left(\bm{\theta}_i - \bm{\mu}_{k+1}\right)^T \bm{\Sigma}_{k+1}^{-1} \left(\bm{\theta}_i - \bm{\mu}_{k+1}\right) \bigg] 
    \\ 
     & \hspace{15pt} + \frac{\xi}{2} \bigg[ 2\epsilon - \text{tr}\left( \bm{\Sigma}_{k+1}^{-1}  \bm{\Sigma}_{k} \right) - n + \log \left(\frac{ \bm{\Sigma}_{k+1}}{ \bm{\Sigma}_{k}} \right) + \left(\bm{\mu}_{k+1} - \bm{\mu}_k \right)^T \bm{\Sigma}_{k+1}^{-1} \left(\bm{\mu}_{k+1} - \bm{\mu}_k\right) \bigg].
\end{align*}
The updates for the mean and covariance can be computed in closed form by setting $\nabla_{\mu_{k+1}}L \coloneqq 0$ and $\nabla_{\Sigma_{k+1}}L \coloneqq 0$, i.e.,
\begin{gather*}
\nabla_{\mu_{t+1}} L = \bm{\Sigma}_{k+1}^{-1} \left[ \sum_{i=1}^{N} \: w_i \: \bm{\theta} + \xi \: \bm{\mu}_{k} - (\xi + 1) \: \bm{\mu}_{k+1} \right] \coloneqq 0 \\
\nabla_{\Sigma_{t+1}} L = \: \bm{\Sigma}_{k+1}^{-1} \: \left( \bm{\Sigma}_{s} + \xi \bm{\Sigma}_{k} +  \xi \left(\bm{\mu}_{k+1} - \bm{\mu}_{k} \right) \left(\bm{\mu}_{k+1} - \bm{\mu}_{k} \right)^{T} - \left(\xi + 1 \right) \: \bm{\Sigma}_{t+1} \right) \: \bm{\Sigma}_{k+1}^{-1}
 \coloneqq 0 
\end{gather*}
The resulting update rule for the mean and covariance is described by
\begin{gather*}
  \bm{\mu}_{k+1} = \frac{\xi^{*} \bm{\mu}_k + \bm{\mu}_s}{1 + \xi^{*}} ,  \hspace{20pt} 
  \bm{\Sigma}_{k+1} = \frac{\bm{\Sigma}_s + \xi^{*} \bm{\Sigma}_k + \xi^{*} \left(\bm{\mu}_{k+1} - \bm{\mu}_k \right)\left(\bm{\mu}_{k+1} - \bm{\mu}_k\right)^T}{1 +\eta^{*}} \\
  \bm{\mu}_s = \sum_{i=0}^{N} w_i \:  \bm{\theta}_i \hspace{20pt} 
  \bm{\Sigma}_s = \sum_{i=0}^{N} w_i \:  \left(\bm{\theta}_i - \bm{\mu}_{k+1}\right) \left(\bm{\theta}_i - \bm{\mu}_{k+1}\right)^T
\end{gather*}
with the optimal multiplier $\xi^{*}$. The optimal multiplier cannot be obtained in closed form as $\bm{\mu}_{k+1}$ and $\bm{\Sigma}_{k+1}$ depend on $\xi$. Hence, $\xi^{*}$ must be obtained by solving
\begin{align*}
    \xi^{*} = \argmax_{\xi} &\bigg[-\log(|\bm{\Sigma}_{t+1}|) - \sum_{i=0}^{N} w_i  \left(\bm{\theta}_i - \bm{\mu}_{k+1}\right)^T \bm{\Sigma}_{k+1}^{-1} \left(\bm{\theta}_i - \bm{\mu}_{k+1}\right) \bigg] 
    \\ 
     + \xi &\bigg[ 2\epsilon - \text{tr}\left( \bm{\Sigma}_{k+1}^{-1}  \bm{\Sigma}_{k} \right) - n + \log \left(\frac{ \bm{\Sigma}_{k+1}}{ \bm{\Sigma}_{k}} \right) + \left(\bm{\mu}_{k+1} - \bm{\mu}_k \right)^T \bm{\Sigma}_{k+1}^{-1} \left(\bm{\mu}_{k+1} - \bm{\mu}_k\right) \bigg]
\end{align*}
with gradient based optimization.

\end{document}